\documentclass{vgtc}                          

\ifpdf
  \pdfoutput=1\relax                   
  \usepackage{graphicx}                
  \DeclareGraphicsExtensions{.pdf,.png,.jpg,.jpeg} 
\else
  \usepackage{graphicx}                
  \DeclareGraphicsExtensions{.eps}     
\fi%

\graphicspath{{figures/}{pictures/}{images/}{./}} 

\usepackage{microtype}                 
\PassOptionsToPackage{warn}{textcomp}  
\usepackage{textcomp}                  
\usepackage{mathptmx}                  
\usepackage{times}                     
\usepackage{cite}                      
\usepackage{tabu}                      
\usepackage{booktabs}                  
\usepackage{subfigure}
\usepackage{booktabs}
\usepackage{adjustbox}
\usepackage{epsfig}
\usepackage{amsmath}
\usepackage{amssymb}
\usepackage[ruled,vlined]{algorithm2e}
\usepackage{caption}
\usepackage{multirow}

\usepackage{hyperref}

\hypersetup{
    pdfproducer={}
}

\onlineid{1065}

\vgtccategory{System}

\vgtcinsertpkg

\title{A Flexible-Frame-Rate Vision-Aided Inertial Object Tracking System \\for Mobile Devices}

\author{Yo-Chung Lau\thanks{\noindent Corresponding author. Email: d06944010@ntu.edu.tw.}\; $^{1,2}$, Kuan-Wei Tseng$^{3}$, I-Ju Hsieh$^{2}$, Hsiao-Ching Tseng$^{2}$, Yi-Ping Hung$^{2}$ \\ \scriptsize $^{1}$Chunghwa Telecom Co., Ltd \quad $^{2}$National Taiwan University \quad $^{3}$Tokyo Institute of Technology  }

\teaser{
  \centering
  \includegraphics[width=\linewidth]{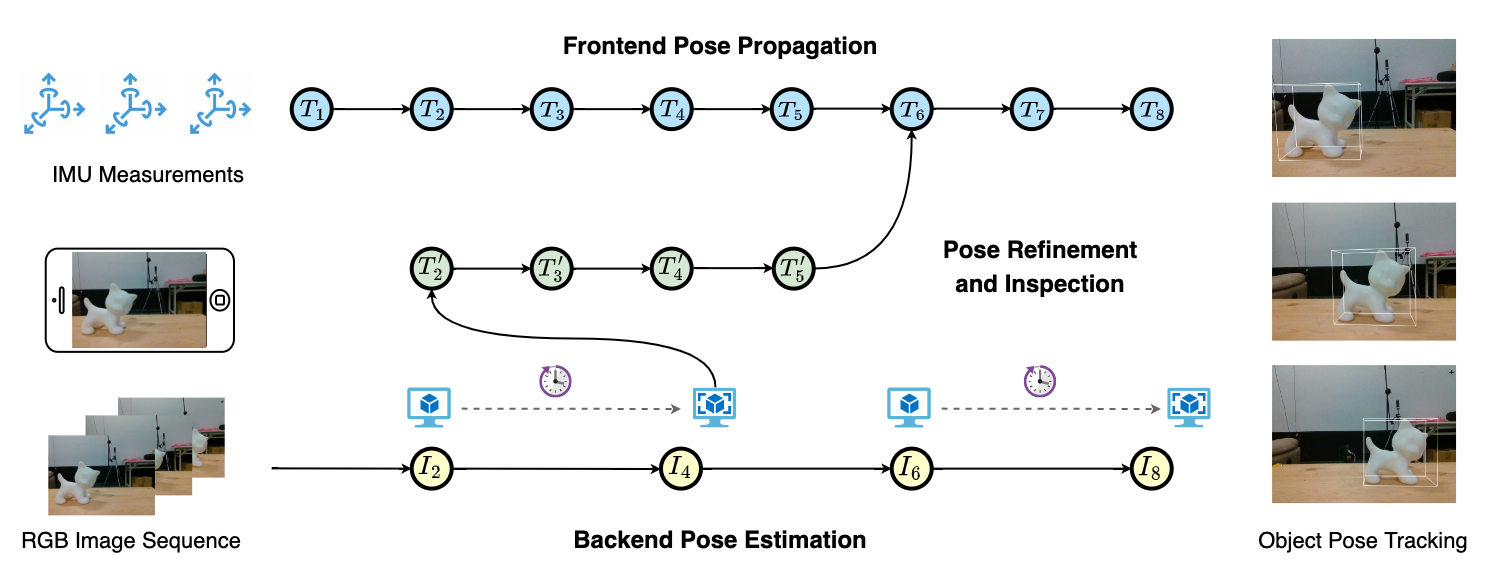}
  \vspace{-5mm}
  \caption{System overview. We propose an object pose tracking system with a client-server architecture for mobile AR applications. The input of the system is IMU measurements and RGB image sequences. On the frontend side (mobile device), we perform the fast pose propagation based on IMU measurements. On the backend side (server), we utilize 6DoF object pose estimation models to estimate the object pose based on RGB images. A more accurate object pose from the backend will be sent back to the frontend to refine the pose and calibrate the biases of IMU measurements. Note that $T_i$ and $I_i$ stand for the object pose and image taken at timestamp $i$, respectively.}
  \label{fig:teaser}
}

\abstract{Real-time object pose estimation and tracking is challenging but essential for emerging augmented reality (AR) applications. In general, state-of-the-art methods address this problem using deep neural networks which indeed yield satisfactory results. Nevertheless, the high computational cost of these methods makes them unsuitable for mobile devices where real-world applications usually take place. In addition, head-mounted displays such as AR glasses require at least 90~FPS to avoid motion sickness, which further complicates the problem. We propose a flexible-frame-rate object pose estimation and tracking system for mobile devices. It is a monocular visual-inertial-based system with a client-server architecture. Inertial measurement unit (IMU) pose propagation is performed on the client side for high speed tracking, and RGB image-based 3D pose estimation is performed on the server side to obtain accurate poses, after which the pose is sent to the client side for visual-inertial fusion, where we propose a bias self-correction mechanism to reduce drift. We also propose a pose inspection algorithm to detect tracking failures and incorrect pose estimation. Connected by high-speed networking, our system supports flexible frame rates up to 120 FPS and guarantees high precision and real-time tracking on low-end devices. Both simulations and real world experiments show that our method achieves accurate and robust object tracking.%
} 

\CCScatlist{
  \CCScatTwelve{Computing methodologies}{Computer vision}{Tracking}{};
  \CCScatTwelve{Human-centered computing}{Mixed/augmented reality}{}{};
}

\begin{document}


\firstsection{Introduction}
\maketitle
The purpose of object pose estimation and tracking is to find the relative 6DoF transformation, including translation and rotation, between the object and the camera. This important task plays a significant role in real-life applications such as adding virtual objects in augmented reality (AR)~\cite{do2010model, su2019deepar} and robotic manipulation~\cite{choi2010real, collet2011moped, tremblay2018deep}. 

Object pose tracking, in contrast to object pose estimation, puts emphasis on tracking object pose in consecutive frames~\cite{hu2019joint, zhong2020seeing}. This is challenging since real-time performance is required to ensure coherent and smooth user experience. Despite the seeming prevalence of solutions, whether they are vision-only~\cite{zhong2020seeing, PoseRBPF} or visual-inertial~\cite{VIPose, Obvit, SchmidtEKF}, such methods are designed to be run on computers or even servers. Hou et al.~\cite{hou2020mobilepose}, based on Sandler et al.~\cite{sandler2018mobilenetv2}, propose lightweight networks to track objects on mobile devices, but hardware requirements are still significant. Moreover, with the development of head-mounted displays, frame rate demands have increased. Although 60~FPS is sufficient for smartphone-based applications, more than 90~FPS is expected for AR glasses to prevent the motion sickness.

\begin{figure*}[t]
  \centering
  \includegraphics[width=\textwidth]{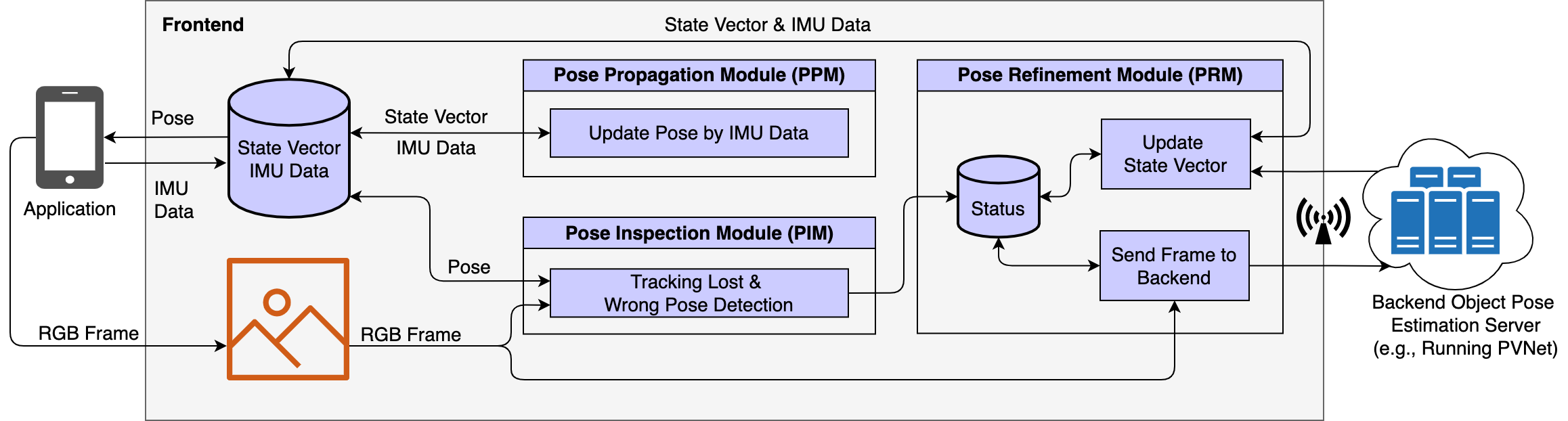}
  \caption{System workflow. The system is composed of a frontend client and a backend server. The frontend performs the fast pose propagation with IMU data and fuses the result of visual pose estimation by the backend server. The state vector contains the device pose and motion information such as the velocity and biases of IMU measurements.}
  \label{fig:systemWorkflow}
\end{figure*}


We thus propose a lightweight system for accurate object pose estimation and tracking with visual-inertial fusion. It uses a client-server architecture that performs fast pose tracking on the client side and accurate pose estimation on the server side. The accumulated error or the drift on the client side is diminished by data exchanges with the server. Specifically, the client is composed of three modules: a pose propagation module (PPM) to calculate a rough pose estimation via inertial measurement unit (IMU) integration; a pose inspection module (PIM) to detect tracking failures, including lost tracking and large pose errors; and a pose refinement module (PRM) to optimize the pose and update the IMU state vector to correct the drift based on the response from the server, which runs state-of-the-art object pose estimation methods using RGB images. This pipeline not only runs in real time but also achieves high frame rates and accurate tracking on low-end mobile devices. The main contributions of our work are summarized as follows:
\begin{itemize}
\item A monocular visual-inertial-based system with a client-server architecture to track objects with flexible frame rates on mid-level or low-level mobile devices.
\item A fast pose inspection algorithm (PIA) to quickly determine the correctness of object pose when tracking.
\item A bias self-correction mechanism (BSCM) to improve pose propagation accuracy.
\item A lightweight object pose dataset with RGB images and IMU measurements to evaluate the quality of object tracking. 
\end{itemize}

\section{Related Work}
\subsection{Object Pose Estimation}
Object pose estimation has long been an open issue; of the many studies on this, some~\cite{wang2019densefusion, liu20203dpvnet, he2021ffb6d} use the depth information to address this problem and indeed yield satisfactory results. Unfortunately, RGB-D images are not always supported or practical in most real use cases. As a result, we then focus on methods that do not rely on the depth information.

\subsubsection{Classical Methods}
Conventional methods which estimate object pose from an RGB image can be classified either as feature-based or template-based. In feature-based  methods~\cite{lowe1999object,rothganger20063d,wagner2008pose}, features in 2D images are extracted and matched with those on the object 3D model. Given the 2D-3D correspondences, the object pose is estimated by solving a PnP problem~\cite{lepetit2009epnp,li2012robust,ferraz2014very}. This kind of method still performs well in occlusion cases, but fails in textureless objects without distinctive features. Template-based  methods~\cite{hinterstoisser2011multimodal, hinterstoisser2012model, ramnath2014car} can handle both textured and textureless objects. Synthetic images rendered around an object 3D model from different camera viewpoints are generated as a template database, and the input image is matched against the templates to find the object pose. However, these methods are sensitive and not robust when objects are occluded.

\subsubsection{Deep Learning-based Methods}
Learning-based methods can also be categorized into direct and PnP-based approaches. Direct approaches regress or infer poses with feed-forward neural networks. SSD6D~\cite{kehl2017ssd}
disentangles the 6D pose into viewpoint and in-plane rotation, first by estimating the rotation and then by inferring the 3D translation with a rotation and bounding box. PoseCNN~\cite{xiang2018posecnn} generates semantic labels and localizes the object center with its distance to the camera via a CNN network. 
PnP-based approaches find 2D-3D correspondences by deep learning, and the estimation of object pose is handled by other PnP solvers. 
PVNet~\cite{peng2019pvnet} selects keypoints by the distance from the center to the surface of the 3D object model. A voting-based algorithm is also used to help find the most correct keypoints in the image, which allows PVNet to effectively tackle occluded objects.
Yu et al.~\cite{Yu2020dfvr} propose differentiable proxy voting loss (DPVL) to reduce the search error of object keypoints. Some studies such as RePOSE~\cite{iwase2021repose} and RNNPose~\cite{xu2022rnnpose} add post-refinement procedures for better pose accuracy. However, these multi-stage pipelines are too slow for real-time applications.

\subsection{Object Pose Tracking}
The purpose of object pose tracking is to estimate object poses in videos. In addition to a single image, temporal information between consecutive frames is also utilized to facilitate estimation. Studies such as Li et al.~\cite{li2018stereo} and Weng et al.~\cite{weng20203d} use a stereo camera or Lidar to help tracking, but this is not practical in real use cases in which only a monocular camera is available. In real-world AR applications, instead of using stereo or RGB-D cameras, IMUs are also commonplace solutions. Thus, we briefly introduce vision-based and visual-inertial-based methods.

\subsubsection{Vision-based Methods}
Classical vision-based methods track features such as SIFT, SURF, and ORB to estimate the correct pose by solving a PnP problem. Likewise, these methods may have high accuracy but their high computational overhead and low robustness to image distortion and self-occlusion are problems~\cite{neumann1999natural}. Based on deep learning, Zhong et al.~\cite{zhong2020seeing} tracks object in video effectively by segmenting objects from the frame even with heavy occlusion.

\subsubsection{Visual-inertial-based Methods}
Conventional visual-inertial fusion using extended Kalman 
filters~\cite{SchmidtEKF, MSCKF} or nonlinear optimization~\cite{Obvit, VINSMono} has
been deployed for AR and robotic applications. However, these suffer from problems of low frame rates and the long delay due to their high computational costs. Recently, 
learning-based methods~\cite{VIPose, VINet, SelectiveFusion} have been proposed which
regress the fused visual and inertial features for camera and object
pose estimation.

\begin{figure*}[t]
  \centering
  \includegraphics[width=\linewidth]{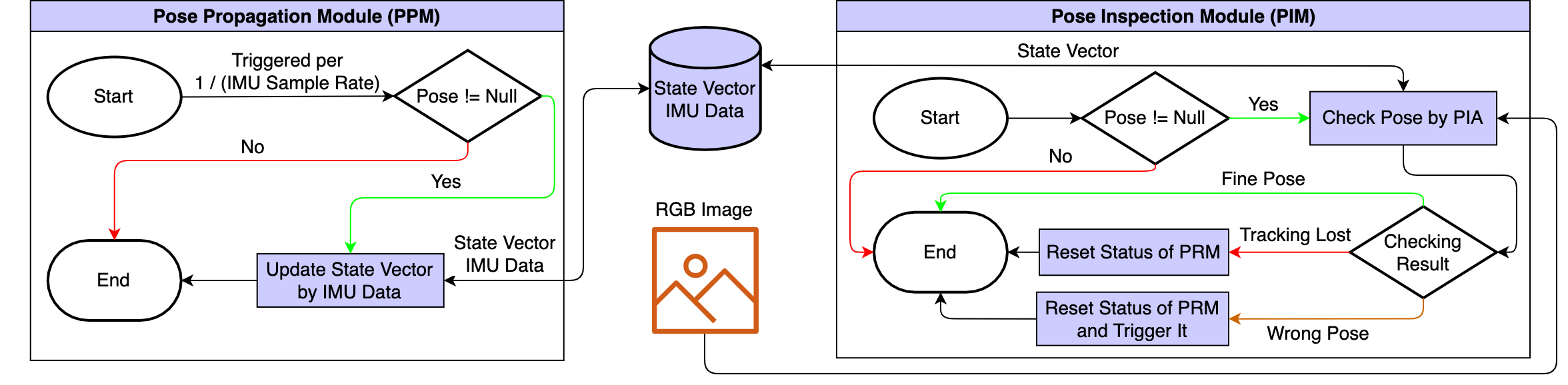}
  \caption{Life diagram of pose propagation module (PPM) and pose inspection module (PIM). PPM updates the client pose by integrating acceleration and angular velocity over time. PIM identifies tracking loss or large pose error by the proposed pose inspection algorithm (PIA).}
  \label{fig:pppim}
\end{figure*}

\section{Proposed Method}
Compared with studies on implementations for PCs or servers, there is a lack of studies
for mobile devices.
MobilePose~\cite{hou2020mobilepose} uses two lightweight neural network models
to track unseen objects and shows strong results on smartphones. It 
achieves 36~FPS on a Galaxy~S20, but the hardware requirements are still 
critical. 
Tracking an object on general mobile devices is difficult,    
but it is necessary for many applications.                                          %
Moreover, to support fluent user interaction, 
applications may have various frame rate requirements (e.g., mixed reality on
smart glasses), which further complicates the task.

\subsection{System Architecture}
\label{section:system-architecture}
To support fluent object tracking on general mid-level or even low-level mobile devices, tasks running on mobile ends should not be overly complicated. We also observe that tracking behavior in real applications is actually a continuous process, and there is typically little motion change between two continuous frames. Thus, based on the object pose in the current frame, the pose in the next frame can be updated just based on the motion change between frames.

We use a client-server design to separate out the frontend and backend tasks. Complicated and time-consuming tasks such as object recognition and precise pose estimation are processed on the backend where powerful computers or servers with high-level hardware are executed. Simple tasks such as pose propagation and checking are handled by the mobile device on the frontend. The two ends are connected to each other through a high-speed network such as 5G or Wi-Fi. In this way, pose updates are rapid, making it possible to achieve real-time tracking with various frame rates.
 
The system workflow is shown in Figure~\ref{fig:systemWorkflow}. 
There are three modules running on the mobile device frontend: the pose propagation module (PPM), the pose inspection module (PIM), and the pose refinement module (PRM). When the system starts, the PRM sends an RGB frame to the backend to obtain the object pose for the frontend initialization. Here, the solution implemented on the backend is not restricted but instead kept open, so that hosts can choose the method that best fits their needs. For example, PVNet \cite{peng2019pvnet} may be a good choice for general use. Once the frontend is initialized, the PPM then regularly updates the pose according to IMU measurements. Meanwhile, the PIM checks the correctness of pose by the pose inspection algorithm (PIA). The PRM repeatedly optimizes the pose computed by the PPM using the response from the backend to maintain the accuracy of tracking.

\subsection{System Modules}
\subsubsection{Pose Propagation Module (PPM)}
The PPM life diagram is shown in Figure~\ref{fig:pppim}. As the PPM is responsible for periodically updating the object pose according to the IMU data by pose propagation, the PPM processing frequency is equal to the IMU sample rate, which is the maximum supportable tracking rate. The IMU data are also saved in the system for later pose updates in the PRM.

\subsubsection{Pose Inspection Module (PIM)}
The PIM checks the correctness of the pose in the current frame. As Figure~\ref{fig:pppim} shows, the PIM uses PIA to examine the pose when a frame arrives, and reports back with \emph{fine pose}, \emph{wrong pose}, or \emph{tracking lost}. The pose is accepted in case of fine pose; other statuses are classified as failures. The PIM reinitializes the PRM status when a failure occurs, and the current frame is processed immediately in the PRM in case of wrong pose.

\subsubsection{Pose Refinement Module (PRM)}
\begin{figure}[t]
  \centering
  \includegraphics[width=\linewidth]{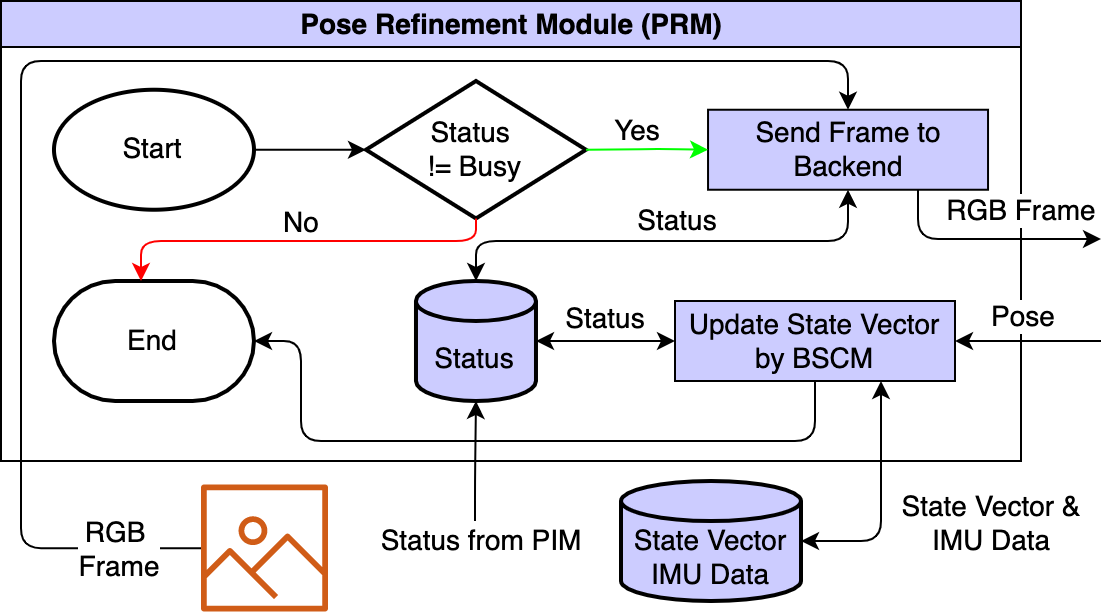}
  \caption{Life diagram of pose refinement module (PRM). PRM refines the tracking pose and correct the accumulated drift. It sends an RGB frame to the backend to query the object pose. Once it gets a response, it updates the system state using the proposed bias self-correction mechanism (BSCM). }
  \label{fig:prm}
  \vspace{-5mm}
\end{figure}

As shown in Figure~\ref{fig:prm}, the PRM attempts to retrieve the frame's correct object pose through the cloud. A pose estimation request with the frame is sent to the backend, after which the module waits for a response. If we assume the sending request and receiving response occurred at time $t_{0}$ and $t_{1}$ respectively, the correct object pose at time $t_{1}$ can be calculated based on the backend result and previously-held IMU data from time $t_{0}$ to $t_{1}$. Meanwhile, we also perform bias self-correction mechanism (BSCM) which leverages the backend correct object pose to compensate for the drift.  In general, the PRM is triggered for every frame, but can also be triggered by the PIM when the wrong pose is found. It is noteworthy the whole processing time should be concerned not to be larger than IMU sample time to ensure the smooth pose updating in the PPM.

\subsection{Pose Propagation by IMU Measurement}
For a static scene, given a known camera pose $M_t$ at time~$t$, we 
have $M_{t+{\delta}t}$ at time $t+\delta t$ with
\begin{equation}
M_{t+\delta t}= M_{t}^{t+\delta t} M_{t}\label{equ:pose_propagation_m},
\end{equation}
where $M_t^{t+{\delta}t} = \left[R_{t+\delta t} \mid T_{t+\delta t}\right]$ is
the camera pose transform matrix. It represents the rotation $R$ and
translation $T$ from time $t$ to $t+\delta t$. When
$\delta t$ is very small, it is safe to calculate $M_t^{t+{\delta}t}$ by
the IMU measurement on devices~\cite{lang2002inertial}.

Theoretically, $R_{t+{\delta}t}$ can be calculated by the change of angular
velocity from the gyroscope, and $T_{t+{\delta}t}$ can be determined by the
moving offset based on the acceleration from the accelerometer as 
follows~\cite{woodman2007introduction}:
\begin{equation}
R_{t+\delta t}=R_{t}\left(I+\frac{\sin \sigma}{\sigma} B+\frac{1-\cos \sigma}{\sigma^{2}} B^{2}\right),
\end{equation}\begin{equation}
B=\left[\begin{array}{ccc}
0 & -\omega_{t+\delta t}^{z} \delta t & \omega_{t+\delta t}^{y} \delta t \\
\omega_{t+\delta t}^{z} \delta t & 0 & -\omega_{t+\delta t}^{x} \delta t \\
-\omega_{t+\delta t}^{y} \delta t & \omega_{t+\delta t}^{x} \delta t & 0
\end{array}\right],
\end{equation}\begin{equation}
\sigma=\left|\omega_{t+\delta t} \delta t\right|,
\end{equation}\begin{equation}
\omega_{t+\delta t}=\left[\omega_{t+\delta t}^{x}, \omega_{t+\delta t}^{y}, \omega_{t+\delta t}^{z}\right]^{T},
\end{equation}
where $I$ is the identity matrix and $\omega_{t+{\delta}t}$ is the device local
angular velocity sampled at time $t+\delta t$.
\begin{equation}
    T_{t+\delta t}=T_{t}+\delta t V_{t+\delta t},
\end{equation}
\begin{equation}
    V_{t+\delta t}=V_{t}+\delta t\left(R_{t+\delta t}\,a_{t+\delta t}-g\right)\label{equ:pose_propagation_v},
\end{equation}
where $V_{t+{\delta}t}$   is the device velocity at time $t+\delta t$,
$a_{t+\delta t}$ is the device local acceleration sampled at time
$t+\delta t$, and $g$ is the acceleration of gravity.

Although it would seem that $M_{t+{\delta}t}$ can be propagated directly from $M_t$ by (\ref{equ:pose_propagation_m})--(\ref{equ:pose_propagation_v}), there may be a problem with correctness. First, IMU data are usually polluted by noise and bias, which makes the pose calculated unreliable. Second, we lack a good initial estimation of system velocity, which results in the translational error. This is non-trivial because there is no reference of the velocity from the backend. The correctness of IMU data and system velocity is so critical to pose propagation that we propose bias self-correction mechanism (BSCM) to compensate for the error. 

\subsection{Bias Self-Correction Mechanism (BSCM)}
\label{section:bscm}
BSCM improves the pose accuracy during pose propagation by removing the bias of IMU data and system velocity. Generally, bias-dependent pose error accumulates and amplifies as the pose is propagated continuously. The estimated rotation $\widehat{R}_{imu}$ obtained from IMU data is defined as
\begin{equation}
\widehat{R}_{imu}=R_{noise}R_{bias} R_{real}\label{equ:rimu},
\end{equation}
where $R_{real}$ is the true rotation, and $R_{bias}$ and $R_{noise}$ are erroneous rotation resulted from the bias and noise. By omitting the noise term, we can approximate the bias $\widehat{R}_{bias}$ using the estimated true rotation $\widehat{R}_{real}$ from the backend as

\begin{equation}
\widehat{R}_{bias}\approx \widehat{R}_{imu}\widehat{R}_{real}^{-1}.
\end{equation}
The corresponding XYZ Euler angles ($\widehat{E}_{bias}$) can be decomposed from
$\widehat{R}_{bias}$, and the time difference ($\Delta t$) from the time of the last triggered backend
pose estimation to now is also known. Hence, the bias of
angular velocity can be written as
\begin{equation}
\widehat{\omega}_{bias}=\frac{\widehat{E}_{bias}}{\Delta t}.
\end{equation}

As for the velocity and acceleration biases, we use the average velocity calculated by consecutive poses as the reference. Assume we have two consecutive frame poses at time $t_{0}$ and $t_{1}$ from the backend, the system average velocity at time $\frac{t_{0} + t_{1}}{2}$ (denoted as $t_0^{1}$) can be represented as
\begin{equation}
\widehat{V}_{avg}=\frac{\widehat{T}_{t_{1}} - \widehat{T}_{t_{0}}}{t_{1} - t_{0}},\label{equ:velocity_avg}
\end{equation}
where $\widehat{T}_{t_{0}}$ and $\widehat{T}_{t_{1}}$ are the translation at time $t_{0}$ and $t_{1}$ estimated by the backend. Thus, we have the system velocity bias at time $t_0^{1}$ as
\begin{equation}
\widehat{V}_{bias}=\widehat{V}_{imu} - \widehat{V}_{avg},
\end{equation}
where $\widehat{V}_{imu}$ is the system velocity closest to time $t_0^{1}$ calculated based on IMU data.
Therefore, the acceleration bias can be derived as
\begin{equation}
\widehat{a}_{bias}=\frac{\widehat{V}_{bias}}{t_{1} - t_{0}}.
\end{equation}

\begin{algorithm}[t]
  \caption{Pose Inspection Algorithm (PIA)}
  \label{alg:the_alg_pia}
  \KwData{\emph{pBox3Ds}: 3D bounding box vertices of object; $K$: camera
  intrinsic parameters
  }
  \KwIn{\emph{poseNow}: current object pose; \emph{poseLast}: object pose in last frame
  }
  \KwOut{\emph{result: finePose, wrongPose, trackingLost}}
  Compute projection points \emph{pBox2DsNow} with \emph{pBox3Ds}, \emph{poseNow}, and $K$\;
  Compute area \emph{areaNow} surrounded by \emph{pBox2DsNow}\;
  \If{(\textit{areaNow} $<$ $\textit{THR}_{area}$)}
  {
    return \emph{trackingLost}\;
  }
  Compute projection points \emph{pBox2DsLast} with \emph{pBox3Ds}, \emph{poseLast}, and $K$\;
  Compute the mean difference \emph{offset} between \emph{pBox2DsNow} with \emph{pBox2DsLast}\;
  \If{(\textit{offset} $\geq$ $\textit{THR}_{2d}$)}
  {
    return \emph{wrongPose}\;
  }
  return \emph{finePose}\;
\end{algorithm}

Through BSCM, we not only estimate and remove the bias of IMU data but also regularly compensate for the velocity.

\subsection{Pose Inspection Algorithm (PIA)}\label{PIA}
A backend pose is reliable and image-independent, but that from the frontend is not always guaranteed. There are two cases in which the propagated pose is unacceptable, including the lost tracking and wrong pose. The tracking is considered lost when the object is out of the camera view or it is too small in the frame. Once the tracking is lost, the pose calculated becomes meaningless and should not be used, even if it is correct. On the other hand, the pose propagation is sensitive to the motion change captured by IMU and may not be reliable when the device moves too rapidly. Hence, for safe use, we also regard the pose as a wrong one if a large motion is detected.

PIA is proposed to quickly check whether the frontend pose is acceptable. The core PIA algorithm is to first find the bounding box of the object in the image based on the pose. We check the projection area and determine for the tracked object whether the area is not less than a threshold, $\emph{THR}_{area}$, here defined empirically as $\frac{\textit{frame area}}{100}$, after which we calculate the mean offset of the vertices from the last frame. If the offset is less than a threshold, $\emph{THR}_{2d}$, we take it as a fine pose; otherwise, we assume the motion is too drastic that the pose propagation might be poor. Thus, the wrong pose is returned. Here, considering that the motion difference between consecutive frames depends on the frame rate, $\emph{THR}_{2d}$ is defined to be inversely proportional to the frame rate with the following relationship:
\begin{equation}
\emph{THR}_{2d}=px_{e}+px_{m} \times \frac{\textit{base rate}}{\textit{frame rate}},
\end{equation}
where $\textit{base rate}$ is defined as 30, and $px_{e}$ and $px_{m}$ are empirically defined as 10, representing reasonable offsets cased by pose error and motion, respectively. We find that if thresholds are set too small, the pose inspection will become too sensitive to the motion change; while there may be an opposite result in cases of larger settings. The PIA processing procedure is represented in the form of the pseudocode in Algorithm ~\ref{alg:the_alg_pia}.

\subsection{System Initialization}\label{PIA}
Since our tracking is based on the client-server data exchanges and pose fusion, to sustain the reliable tracking by pose propagation on the frontend, we need to have appropriate system state vectors of the initial pose, velocity, and data bias, which are estimated through BSCM. In fact, for some data such as the IMU bias information, it is tricky to predict them in advance to help the system initialization. For instance, before tracking, we can let the mobile device static and collect the IMU measurements to gauge the initial data bias.


\section{Experiments}
Since there is no publicly available dataset for visual-inertial object tracking, we first evaluated the proposed system with a simulated dataset using Gazebo (v11.0.0)~\cite{gazebo}. Afterwards, we verified our method with real world data.

\begin{figure}[t]
    \centering
    \subfigure[Simulated Data]{
        \includegraphics[width=0.47\linewidth, height=3cm]{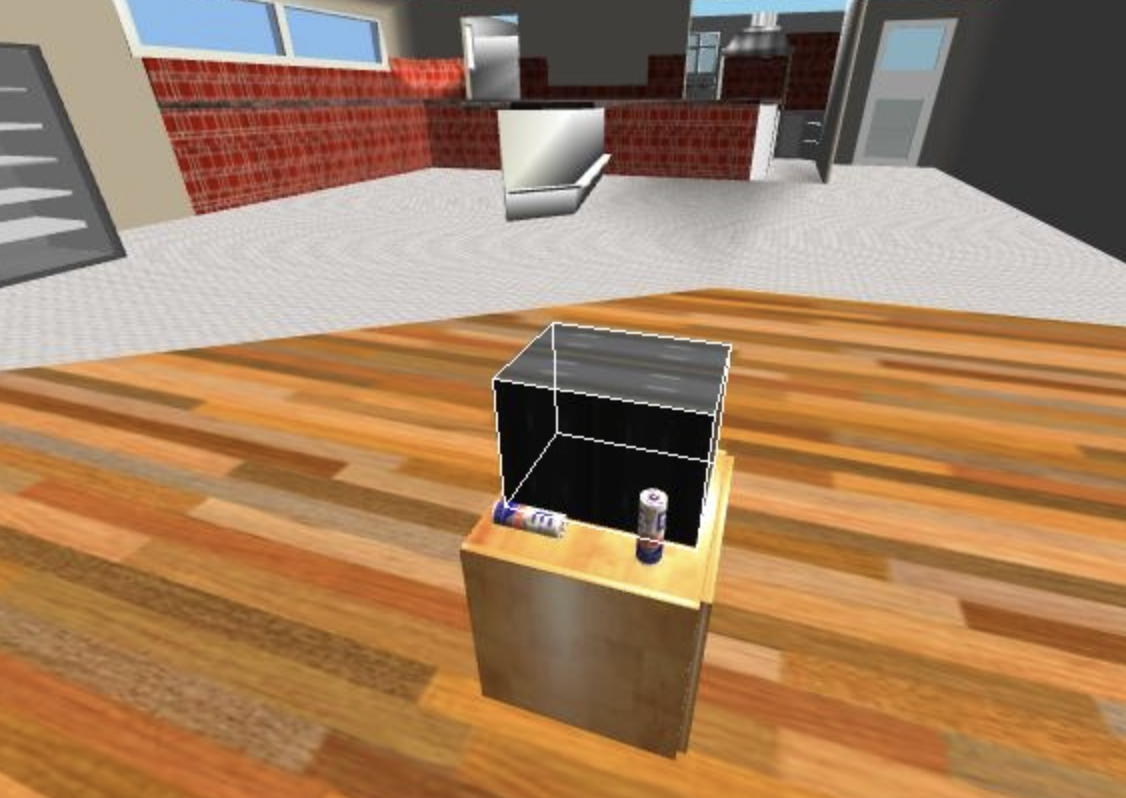}
        \label{fig:scene1}
    }
    \subfigure[Real World Data]{
        \includegraphics[width=0.47\linewidth, height=3cm, trim={0cm 0cm 0cm 0.2cm}, clip]{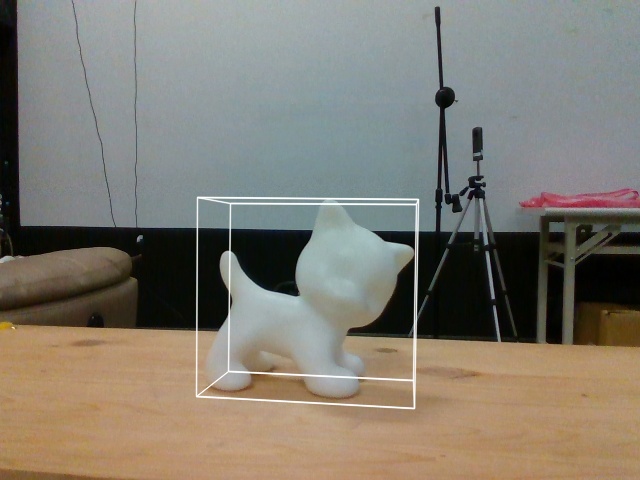}
        \label{fig:real-cat}
    }
    \vspace{-3mm}
    \caption{To evaluate our model, we construct two datasets, including a simulated one in Gazebo physical engine and a real world one in a room with Vicon motion capture system.}
    \vspace{-3mm}
    \label{fig:scenes}  
\end{figure}

\begin{table}[t]
\caption{Simulation specifications}
\label{table:spec}
\begin{adjustbox}{width=\columnwidth,center}
\centering
\begin{tabular}{@{}cccccc@{}}
\toprule
\multicolumn{2}{c}{Quantity} & \multicolumn{2}{c}{Translational} & \multicolumn{2}{c}{Angular} \\ \midrule
\multicolumn{2}{c}{Script} &
  \begin{tabular}[c]{@{}c@{}}Average \\ speed \\($m/s$) \end{tabular} &
  \begin{tabular}[c]{@{}c@{}}Acceleration \\ range \\($m/s^2$) \end{tabular} &
  \begin{tabular}[c]{@{}c@{}}Average \\ rate \\($\mathit{rad}/s$)\end{tabular} &
  \begin{tabular}[c]{@{}c@{}}Acceleration \\ range \\($\mathit{rad}/s^2$)\end{tabular} \\ \midrule
                 & Easy      & 0.062     & [0,0.191)   &  0.001        & 0                \\ \cmidrule(l){2-6} 
Translational    & Medium    & 0.123     & [0.173,0.364) &  0.014        & [0,0.152)         \\ \cmidrule(l){2-6} 
                 & Hard      & 0.182     & [0.346,0.537) &  0.041        & [0.151,0.303)               \\ \midrule
                 & Easy      & 0.073     & [0.010,0.016) &  0.056    &[0,0.038) \\ \cmidrule(l){2-6} 
Circular         & Medium    & 0.147     & [0.039,0.063) &  0.330  &[0,0.150)\\ \cmidrule(l){2-6} 
                 & Hard      & 0.229     & [0.088,0.140) &  0.402   &[0,0.338)\\ \bottomrule
\end{tabular}
\end{adjustbox}
\end{table}

\subsection{Experiments on Simulated Data}
As Figure~\ref{fig:scene1} shows, we created an indoor scene in which the target object was placed on a table and its initial distance to the smartphone with an on-board monocular camera and IMU was about 1.2 meters. The camera captured images up to 120~FPS, and the IMU sample rate was about 200~Hz throughout the experiment. The IMU noise parameters were 6.63$\times$10$^{-5}$ ${\mathit{rps}}/\sqrt{\mathit{Hz}}$ for the gyroscope and 7.35$\times$10$^{-4}$ $\,m/s^2\sqrt{\mathit{Hz}}$ for the accelerometer, which resembled the LSM6DSM, a consumer-level IMU chipset in the Google Pixel 2. In our simulations, the camera followed the target object in the scene, and its movement followed predefined motion programs. We then compared the tracking results with the true poses for evaluation.

\subsubsection{Motion Program}
To simulate the diverse interaction behaviors of real AR applications, we used translational and circular motion scripts, each with three levels of difficulty. The script details are shown in Table~\ref{table:spec}. Each script ran for 30 seconds. For the translational motion, we primarily moved the camera using a combination of dolly, pedestal, and trucking; for the circular motion, we moved and rotated the camera around the object. We also applied random additional force and torque to the camera, roughly resembling real-world vibration or shaking. Thus, the higher the difficulty, the faster the movement and the larger the force and torque.

\subsubsection{Networking and Transmission}
\label{section:networking and transmission}
According to the minimum standards of a 5G network~\cite{alliance20155g}, we set the network transmission speed and data propagation delay to 50~Mbps and 10~ms respectively. Also, as real-world transmission is imperfect, we added a uniformly distributed extra delay of 0--30~ms. The size of the transmitted data was 100 kB, including a 90\;kB 640$\times$480 JPEG image with a compression ratio of 10, and 10~kB of metadata containing the state vector and other information. Once the server received the data, the backend task started instantly without delay. The backend response time was thereby

\begin{equation}
    \begin{aligned}
     t &= \frac{100\;(\text{data size}) \times 8\; (\text{kB to kb})}{1024\; (\text{kb to Mb}) \times 50\; (\text{network speed})} \times 10^3\;(\text{s to ms})\\ &\quad + 10\; (\text{propagation delay}) \times 2\; (\text{back and forth}) \\&\quad + U\;(0,30)\;  (\text{extra delay}) \approx (35,66) \text{ ms}   
    \end{aligned}
\end{equation}

\begin{figure}[t]
  \centering
  \includegraphics[width=\linewidth, trim={0cm 0cm 0cm 0.5cm}]{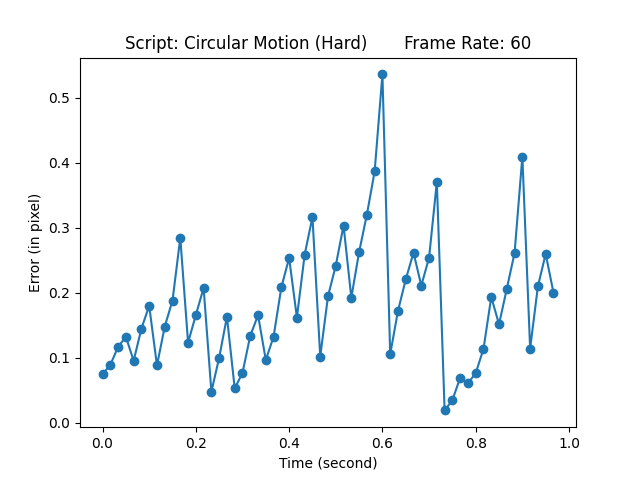}
  \caption{2D projection errors among the pose refinement processes. The tracking error in our method will be fixed periodically by BSCM with the backend responses. Each local fluctuation here (between the local maximum and the minimum) can be considered as a cycle of the pose propagation and refinement.}
  \label{fig:2derror1s}
\end{figure}

\begin{table*}[]
\caption{Comparison of different backend configuration. Two backend settings (GT: ground-truth, Noisy: ground-truth contaminated with Gaussian noise) are compared. We report mean position (mm), orientation (degree), and 2D projection error (pixel), respectively.}
\label{table:poseerrors-gazebo}
\centering
\begin{tabular}{cccccccc}
\toprule
\multicolumn{2}{c}{Motion Type}    & \multicolumn{3}{c}{Trasnlational}                          & \multicolumn{3}{c}{Circular}                     \\ \midrule
Backend                & Frame rate & Easy                    & Medium          & Hard           & Easy           & Medium         & Hard           \\ \midrule
\multirow{4}{*}[-6pt]{GT}    & 30 FPS    & 1.17/0.00/0.20          & 2.29/0.00/0.45  & 3.39/0.01/0.69 & 0.21/0.02/0.10 & 0.50/0.03/0.17 & 1.02/0.06/0.32 \\ \cmidrule(l){2-8}
                       & 60 FPS    & 1.05/0.00/0.18          & 2.07/0.00/0.41 & 3.13/0.01/0.64 & 0.19/0.02/0.08 & 0.49/0.03/0.15 & 0.90/0.05/0.25 \\ \cmidrule(l){2-8}
                       & 90 FPS    & 1.04/0.00/0.18          & 2.02/0.00/0.40  & 2.99/0.01/0.60 & 0.21/0.02/0.09 & 0.51/0.03/0.18 & 0.94/0.05/0.29 \\ \cmidrule(l){2-8}
                       & 120 FPS   & 1.02/0.00/0.18          & 1.97/0.00/0.39 & 2.93/0.01/0.59 & 0.22/0.02/0.10 & 0.53/0.04/0.19 & 0.98/0.06/0.32 \\ \midrule
\multirow{4}{*}[-6pt]{Noisy} & 30 FPS     & 4.89/0.47/1.96          & 5.16/0.48/2.10  & 5.78/0.47/2.15 & 4.68/0.48/2.00 & 4.73/0.47/2.00 & 4.80/0.48/2.01 \\ \cmidrule(l){2-8}
                       & 60 FPS     & 5.16/0.50/2.09          & 5.39/0.50/2.15  & 5.86/0.50/2.28 & 5.03/0.50/2.09 & 5.04/0.51/2.14 & 5.12/0.51/2.10 \\ \cmidrule(l){2-8}
                       & 90 FPS     & 5.24/0.52/2.15          & 5.53/0.51/2.22  & 5.92/0.52/2.31 & 5.13/0.51/2.17 & 5.12/0.51/2.14 & 5.21/0.52/2.16 \\ \cmidrule(l){2-8}
                       & 120 FPS    & 5.26/0.51/2.14          & 5.52/0.52/2.25  & 5.98/0.52/2.33 & 5.16/0.52/2.19 & 5.27/0.52/2.16 & 5.26/0.53/2.21 \\ \bottomrule
\end{tabular}
\end{table*}

\subsubsection{Computational Power}
Mid-level or low-level phones released 3~to 5 years ago, such as the Google Pixel 2 released in 2017, generally have limited computational power. Thus, we restricted the CPU performance of our testing device so that its computational power was commensurate with the Google Pixel 2. We referred to the CPU multi-core scores from Geekbench~\cite{geekbenchGooglePixel2}. The performance ratio of two CPUs can be directly obtained from the ratio of multi-core scores. Based on this assumption, we delayed the processing time of each operation on the frontend by adding extra sleep. Specifically, the multi-core scores of our device and Google Pixel 2 were 3185 and 1294, respectively. The performance ratio was thereby \mbox{3185 / 1294} $\approx$ 2.46. Hence, we extended the processing time of each frontend operation by 246\%.

\begin{figure*}[t]
    \centering
    \subfigure[Simulated Data (Backend:GT)]{
        \includegraphics[width=0.32\linewidth, trim={0.8cm 0cm 0.5cm 0.2cm}]{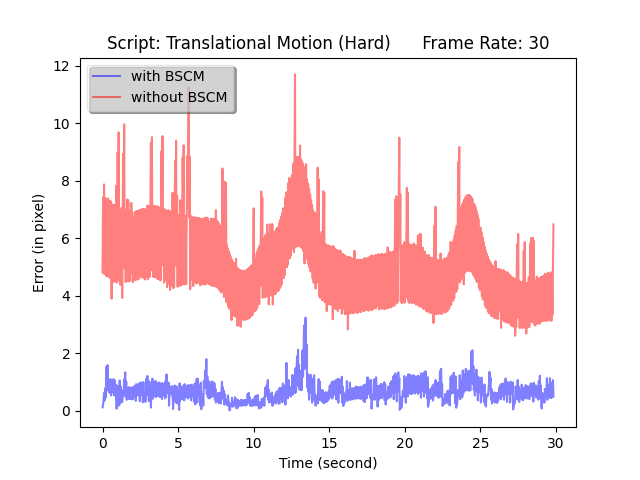}
        \label{fig:woBSCM-pure}
    }
    \subfigure[Simulated Data (Backend:Noisy)]{
        \includegraphics[width=0.32\linewidth, trim={0.8cm 0cm 0.5cm 0.2cm}, clip]{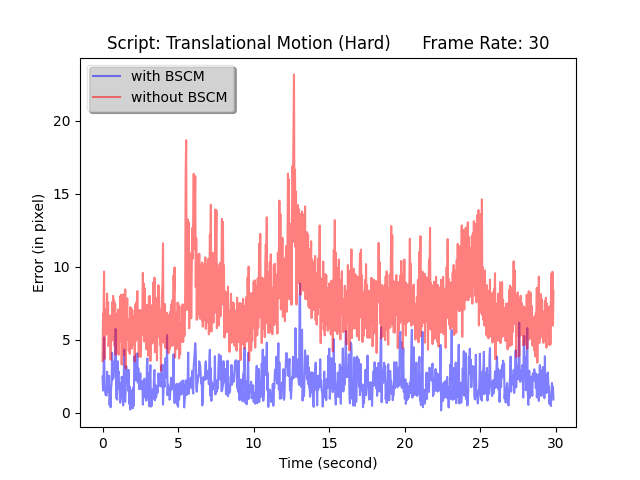}
        \label{fig:woBSCM-noised}
    }
    \subfigure[Real World Data]{
        \includegraphics[width=0.32\linewidth, trim={0.2cm 0cm 1.1cm 0.2cm}, clip]{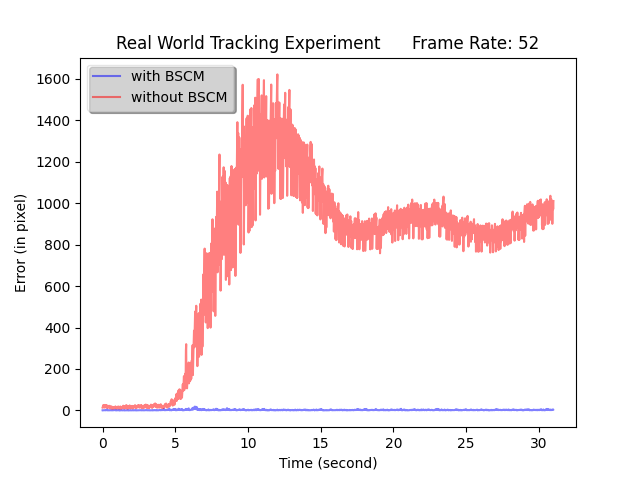}
        \label{fig:woBSCM-real}
    }
    \caption{Ablation study on Bias Self-Correction Mechanism (BSCM) by comparing the 2D projection error on the simulated and real world data. Similar to Table 2, we present the results of two backend settings on the simulated data. The primary difference between the simulated and the real data lies in the larger and more unpredictable IMU biases in the real world.}
    \label{fig:bscm_ablation}  
\end{figure*}

\subsubsection{Simulation Results and Discussion}
We reported the mean 6DoF pose error (translation and rotation) and 2D projection error of our tracking results in Table~\ref{table:poseerrors-gazebo}. For the qualitative visualization of our tracking system, please refer to the supplementary video. Following the convention, we computed the 2D projection error by measuring the distance projected on the image plane using the estimated pose and the true pose. To understand the influence of the accuracy of the pose responded from the backend, there were two kinds of backend responses in our simulations, including the ground-truth (GT) and noisy poses. Noisy poses were generated by GT ones with Gaussian noise.

The error increased in harder sequences with higher motion complexity. The tracking was very smooth and accurate in all experiments if GT backend was utilized. Its 2D projection error was less than 1 pixel among all sequences regardless of motion complexity. The result has validated our core element that once the true pose is received from the backend, the pose is refined and the error should decrease. More clearly, Figure~\ref{fig:2derror1s} shows the object 2D projection error of frames (points) within a short period of time in tracking. Each fluctuation can be taken as one process of the pose refinement. The pose error first kept increasing to the peak, after which the refinement occurred, leading to fixing the pose error of the next frame. Specifically, for each fluctuation, the pose refinement would happen in the interval between the peak and foot.

On the other hand, if the backend returned a noisy measurement, which reflected the practical situation, then there was an admitted drop in the performance. The average 2D projection error became about 2 pixels, but was still below the critical threshold of 5 pixels. This demonstrated that even with imperfect backend solution, our system was still capable of maintaining acceptable poses and reducing the impact of biases.

Moreover, we analyzed the influence of frame rate on tracking accuracy. If GT backend was used, the error decreased at higher frame rate due to shorter waiting time when the backend server was idle. As a consequence, it benefited the pose refinement and pose propagation afterwards. Nevertheless, a counterproductive result would happen in noisy backend case. In other words, if the backend pose was noisy, then higher frame rate would lead to higher error. We believe that the reason behind this contradictory phenomenon was that the error accumulated during the latency was less than the noisy backend poses. In this case, we actually added additional error to the system when there was no serious drift on the frontend. 

Besides accuracy, we also studied the execution speed of our tracking system. Our systems can support high frame rate up to 120 FPS. The average pose inspecting time (PIM) and updating time (PRM) on the frontend were around 0.35~ms and 0.86~ms after the CPU performance conversion. The frontend processing is as fast as expected and would not influence the regular PPM pose updating. 

To verify the effectiveness of the proposed pose refinement mechanism, we conducted ablation studies on PRM by disabling the BSCM and the PIA. The results are shown in Figure~\ref{fig:woBSCM-pure} and Figure~\ref{fig:woBSCM-noised}. Similarly, we presented two backend settings. Note that once the BSCM and PIA were disabled, the IMU biases would never be removed during pose propagation. We found that the error could still decrease because we would still re-propagate the pose if there was a response from the backend. However, the error was more than 2 times larger compared to that of full model. In addition, if we remove the entire backend system, i.e., making the system work only on pose propagation itself, the error would be so large that the system would never have any practical utility.

\begin{table}[t]
\centering
\caption{The mean pose error of evaluation on the real world data. There are five 30-seconds videos whose camera movements follow the common interaction of AR applications.}
\begin{tabular}{@{}llllll@{}}
\toprule
                                                                       & \#1  & \#2  & \#3  & \#4  & \#5  \\ \midrule
Position Error (mm)                                                          & 1.65 & 2.06 & 2.01 & 1.25 & 1.30 \\ \midrule
Orientation Error (degree)                                                         & 1.54 & 6.96 & 8.17 & 8.23 & 3.91 \\ \midrule
\begin{tabular}[c]{@{}l@{}}2D Projection Error (pixel)\end{tabular} & 2.56 & 2.93 & 3.37 & 2.21 & 2.30 \\ \bottomrule
\end{tabular}
\label{table:poseerrors-real}
\end{table}

\subsection{Experiments on Real World Data}
\subsubsection{Real World Data Collection}
Besides simulation, we collected our own data to validate our system under the real world scenario. We utilized Intel Realsense D435i camera to capture RGB images, depth images, and IMU measurements in a room with Vicon motion capture system. The average frame rate of RGB images was 52 FPS, while the IMU sampling rate was 200 Hz. The target object was a 3D printed cat from LineMOD~\cite{hinterstoisser2012model}. As shown in Figure~\ref{fig:real-cat}, we put the cat model on the table and moved the camera in a common movement pattern of general interactive AR applications. 

To obtain the object pose with respect to the camera is a non-trivial work. Inspired by \cite{Laval}, we first calibrated the extrinsic parameters using a checkerboard with Vicon markers attached at each corner. Combined with intrinsic parameters, we then had the transformation between image plane and the global coordinate system defined by Vicon. Next, we back-projected depth maps with corresponding camera poses to reconstruct a 3D point cloud model of the cat under the global coordinate system. We then registered the reconstructed 3D point cloud model with the original CAD model of the cat using Iterative Closest Points (ICP) to obtain the transformation between the predefined object-centric coordinate system and the global coordinate system. Eventually, we could compute the 6DoF object pose, which was the transformation between object-centric coordinate system and camera coordinate system. In practice, since the depth maps may fail to precisely segment the silhouette of the object, we performed some post processing with ICP by each frame. Although the object poses are still noisy, it serves a great opportunity to examine whether our system would work in the real world using where the backend would return noisy poses. 




\subsubsection{Real World Experiment Results and Discussion}
We summarized the mean pose error and 2D projection error of each sequence in Table~\ref{table:poseerrors-real}. We found that compared to the simulated data, the rotational error was large. This could be attributed to the shaking of the capturing device during data collection. Videos shot by handheld devices without stabilizer would inevitably contain such movements. As a consequence, the average angular velocity (and acceleration) was larger than the simulated data. Besides, the magnitude of translational error was about that of simulated ones with GT backend although it utilized noisy backend poses. This resulted from the slower moving speed in real world applications. We moved the camera faster in the simulated data to examine the robustness of model. However, in real world AR applications, it is less possible that users would move so fast. Thus, the smaller error was due to the slower movement.

Similar to simulation, Figure~\ref{fig:woBSCM-real} shows the performance of real world tracking experiments with/without BSCM. As the results we saw in simulation, the real world tracking with BSCM still performed well with a good pose accuracy. In addition, we also found the huge 2D projection and pose errors when BSCM was not implemented in tracking, while the problem was not so serious in simulation. We blamed a larger noise of IMU data in real world tracking for this phenomenon. In fact, the noise in IMU measurement is related to many factors, including the type of movements, the elapsed time of using, the device temperature, and so on, which may not be totally simulated and reflected by Gazebo. The influence of the noise in IMU data should be taken into consideration seriously, and our BSCM indeed was again proven its importance and effectiveness to keep accurate poses in tracking.

\section{Conclusions}
We review the importance of high speed object pose estimation and tracking using only RGB images taken by mobile devices. To this end, we present a flexible-frame-rate vision-aided inertial object tracking solution with low computational overhead. The client server architecture allows us to reduce the computational cost on the frontend while achieving high tracking accuracy by using errorless poses computed by backend servers. To ensure robustness, we devise pose inspection algorithm (PIA) to quickly examine the reliability of the object pose. Most important of all, we propose a bias self-correction mechanism (BSCM) to alleviate the error that accumulates over time. We not only formulate our method mathematically, but also verify its feasibility via simulations and real world experiments. We believe that this research and the data we collected for visual-inertial object pose estimation and tracking will facilitate the future development of AR applications.


\acknowledgments{
This work was partially supported by Ministry of Science and Technology in Taiwan (MOST 110-2218-E-002-033-MBK and 109-2221-E-002-207-MY3).}

\bibliographystyle{abbrv-doi}

\bibliography{main}
\end{document}